%% file: root.tex
\documentclass[letterpaper, 10 pt, conference]{IEEEtran}
\IEEEoverridecommandlockouts                                       
 
\usepackage{cite}
\usepackage{algorithm, algpseudocode}  
\usepackage{graphics} 
\usepackage{epsfig} 
\usepackage{amssymb} 
\usepackage{amsmath} 
\usepackage{latexsym}
\usepackage{mathrsfs} %
\usepackage{xcolor}
\usepackage[normalem]{ulem}
\usepackage{bm}       
\usepackage{graphicx}
\usepackage{verbatim}
\usepackage{balance}
\usepackage{geometry}

\ifCLASSOPTIONcompsoc
    \usepackage[caption=false, font=normalsize, labelfont=sf, textfont=sf]{subfig}
\else
\usepackage[caption=false, font=footnotesize]{subfig}
\fi

\usepackage[linkbordercolor={1 1 1},citebordercolor={1 1
  1},urlbordercolor={0.0 0.0
  0.0},urlcolor=blue,colorlinks=true,linkcolor=black,citecolor=black]{hyperref}
\usepackage{url}

\newcommand{\algo}{{\sc\textsf{CoSpar}}}
\newcommand{\newsec}[1]{\vspace{2mm} \noindent \textbf{#1.} }
\newcommand{\newsubsec}[1]{\vspace{2mm} \noindent \underline{#1.} }

\title{\vspace{10pt}\LARGE \bf
Preference-Based Learning for Exoskeleton Gait Optimization}

\author{Maegan Tucker$^{*1}$, Ellen Novoseller$^{*2}$, Claudia Kann$^{1}$, \\ Yanan Sui$^{3}$, Yisong Yue$^{2}$, Joel W. Burdick$^{1, 2}$, and Aaron D. Ames$^{1, 2}$
\thanks{*These two authors contributed equally to this work.}
\thanks{This research was supported by NIH grant EB007615, NSF NRI award 1724464, NSF Graduate Research Fellowship No. DGE‐1745301, and the Caltech Big Ideas and ZEITLIN Funds.}
\thanks{This work was conducted under IRB No. 16-0693.}
\thanks{$^{1}$Authors are with the Department
of Mechanical and Civil Engineering, California Institute of Technology,
Pasadena, CA 91125. \texttt{[mtucker,ckann,jburdick,ames]@caltech.edu}}
\thanks{$^{2}$Authors are with the Department
of Computing and Mathematical Sciences, California Institute of Technology,
Pasadena, CA 91125. \texttt{[enovoseller,yyue]@caltech.edu}}
\thanks{$^{3}$Author is with the School of Aerospace Engineering, Tsinghua University, Beijing, China 100084. \texttt{ysui@tsinghua.edu.cn}}%
}

\begin{document}

\maketitle
\thispagestyle{empty}
\pagestyle{empty}

\begin{abstract}

This paper presents a personalized gait optimization framework for lower-body exoskeletons. Rather than optimizing numerical objectives such as the mechanical cost of transport, our approach directly learns from user preferences, e.g., for comfort. Building upon work in preference-based interactive learning, we present the \algo~algorithm. \algo~prompts the user to give pairwise preferences between trials and suggest improvements; as exoskeleton walking is a non-intuitive behavior, users can provide preferences more easily and reliably than numerical feedback. We show that \algo~performs competitively in simulation and demonstrate a prototype implementation of \algo~on a lower-body exoskeleton to optimize human walking trajectory features. In the experiments, \algo~consistently found user-preferred parameters of the exoskeleton's walking gait, which suggests that it is a promising starting point for adapting and personalizing exoskeletons (or other assistive devices) to individual users.
\end{abstract}


\input{./Sections/Introduction.tex}            
\input{./Sections/GaitGeneration.tex}            
\input{./Sections/Algorithm.tex}      
\input{./Sections/Simulation.tex}                  
\input{./Sections/Experiments.tex}                 
\input{./Sections/Conclusion.tex}              

\vspace{6mm}
\section*{ACKNOWLEDGMENTS}
The authors would like to thank the volunteers who participated in the experiments, as well as the entire Wandercraft team that designed Atalante and continues to provide technical support for this project.

\newpage
\bibliographystyle{IEEEtran}
\balance
\bibliography{IEEEabrv,References}

\end{document}

%% file: Sections/Introduction.tex
\section{INTRODUCTION}
\label{sec:intro}
The field of human-robot interaction is receiving increasing attention in many application domains, from mobility assistance to autonomous driving, and from education to dialog systems. In many such domains, for a robotic system to interact optimally with a human user, it must adapt to user feedback. In particular, learning from user feedback could help to improve robotic assistive devices.

This work focuses on optimizing walking gaits for a lower-body exoskeleton, Atalante, to maximize user comfort. Atalante, developed by Wandercraft \cite{wandercraft}, uses 12 actuated joints to restore mobility to individuals with lower-limb mobility impairments, which could potentially benefit approximately 6.4 million people in the United States alone \cite{HerrActiveOrthoses}. Existing work with Atalante has demonstrated dynamically-stable walking using the method of partial hybrid zero dynamics (PHZD), originally designed for bipedal robots \cite{harib2018feedback,gurriet2018towards, agrawal2017first}. While this method generates stable bipedal locomotion, there is no current framework to optimize for  comfort; yet, user comfort should be a critical objective of gait optimization for exoskeleton walking. While existing methods \cite{ames2014human} can generate human-like walking gaits for bipedal robots, it is unlikely that these methods fulfill the preferences of individuals using robotic assistance.

Existing human-in-the-loop algorithms optimize quantitative metrics such as metabolic expenditure \cite{zhang2017human}; however, since the goal of this work is to optimize for user comfort, the presented learning approach uses user preferences obtained from sequential gait trials. By directly incorporating personalized feedback, we avoid making overly-strong assumptions about gait preference, or optimizing for a numerical quantity not aligned to personalized comfort.

For exoskeleton gait generation, as in many real-world settings involving people \cite{amodei2016concrete, argall2009survey, basu2017you}, it is challenging for people to reliably specify numerical scores or provide demonstrations. In such cases, the users' \textit{relative preferences} measure their comfort more reliably. Previous studies have found preferences to be more reliable than numerical scores in a range of domains, including information retrieval \cite{chapelle2012large} and autonomous driving \cite{basu2017you}.

\begin{figure}[t]
    \centering
    \includegraphics[width = 0.72\linewidth]{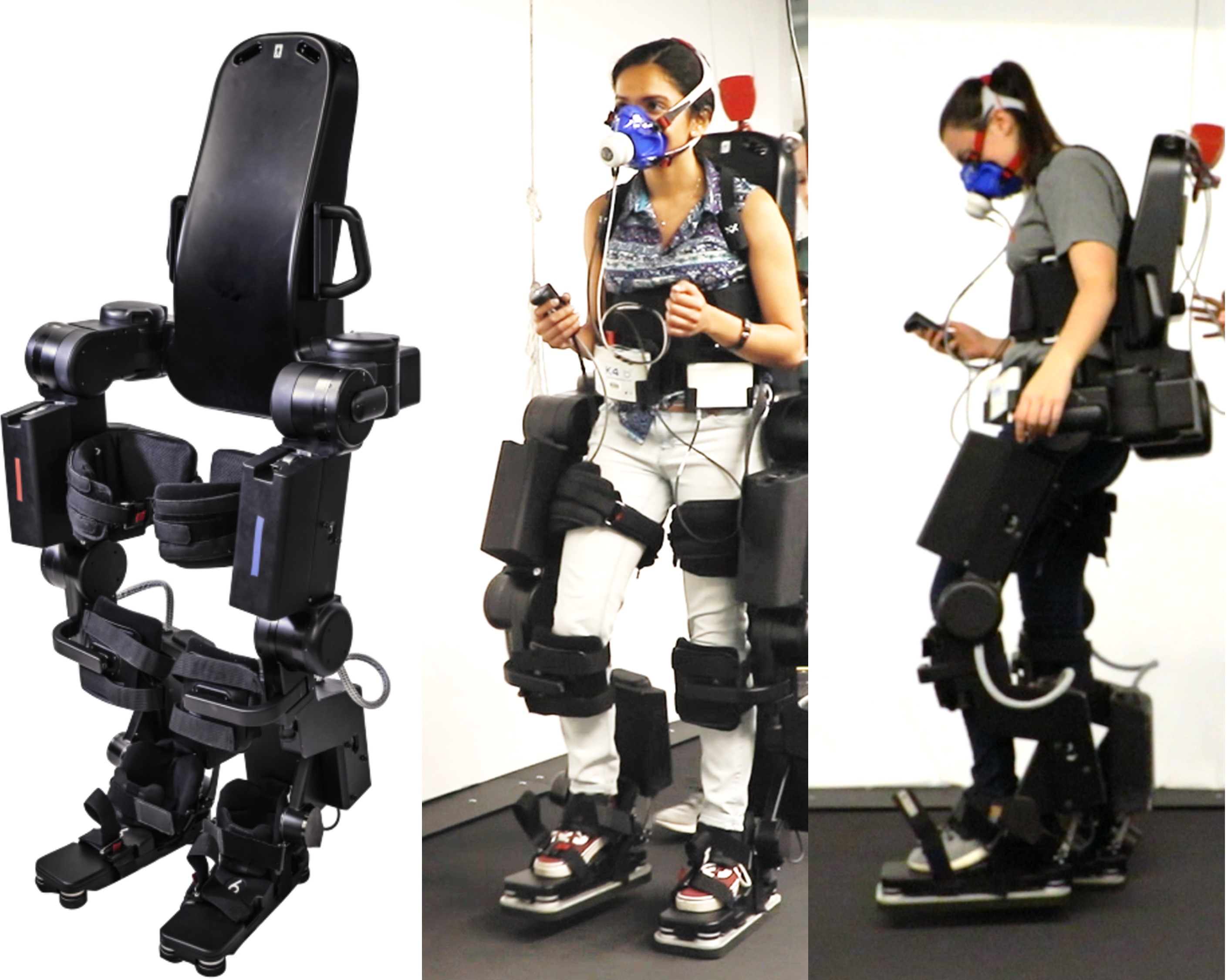}
    \caption{Atalante Exoskeleton with and without the user. The user is wearing a mask to measure metabolic expenditure.}
    \label{fig:atalante}
    \vspace{22.5pt}
\end{figure}

Building upon techniques from  dueling bandits \cite{sui2017multi, sui2018advancements, yue2012k} and coactive learning \cite{shivaswamy2012online, shivaswamy2015coactive}, we propose the \algo~algorithm to learn user-preferred exoskeleton gaits. \algo~is a mixed-initiative approach, which both queries the user for preferences and allows the user to suggest improvements. 
We also validate \algo~in simulation and human experiments, in which \algo~finds user-preferred gaits within a gait library. This procedure not only identifies users' preferred walking trajectories, but also provides insights into the users' preferences for certain gaits.

%% file: Sections/GaitGeneration.tex
\section{Gait Generation for Bipedal Robots}\label{sec:stable_gaits}
Many existing lower-body exoskeletons either require the use of arm-crutches \cite{Ekso,rewalk,indego} or use slow static gaits with speeds around 0.05 m/s \cite{rex}. Using the PHZD method, dynamic crutchless exoskeleton walking has been demonstrated to generate dynamically-stable gaits. We briefly explain this method to illustrate how it can be adapted based on user preferences; for more details, refer to \cite{harib2018feedback, gurriet2018towards, agrawal2017first}.

\newsec{Partial Hybrid Zero Dynamics Method} Systems with impulse effects, such as ground impacts, can be represented as \textit{hybrid control systems} \cite{westervelt2003hybrid, bainov1989systems,ye1998stability}. Summarizing from \cite{gurriet2018towards}, the natural system dynamics can then be represented on an invariant reduced-dimensional surface, termed the \textit{zero dynamics} surface \cite{westervelt2018feedback}, by appropriately defining the \textit{virtual constraints} and using a feedback-linearizing controller to drive them to zero. Since the exoskeleton's desired forward hip velocity is constant and its actual velocity experiences a jump at impact, the \textit{partial zero dynamics} surface is considered. The \textit{virtual constraints} are defined as the difference between the actual and desired outputs:
\begin{align}
    y_1(q,\dot{q}) &= y_1^{a}(q,\dot{q}) - v_d \label{eq: virt1}\\ 
    y_2(q, \alpha) &= y_2^{a}(q) - y_2^{d}(\tau(q), \alpha) \label{eq: virt2},
\end{align}
where the actual outputs $y_1^a$ and $y_2^a$ are velocity-regulating and position-modulating terms, respectively. The output $y_1^a$ is driven to a constant desired velocity $v_d$, while $y_2^a$ is driven to a vector of desired trajectories, $y_2^d$. The trajectories $y_2^d$ are represented using a B\'{e}zier polynomial with coefficients $\alpha$ and state-based timing variable $\tau(q)$. 

According to Theorem 2 in \cite{ames2014human}, if there exist virtual constraints that yield an impact-invariant periodic orbit on the \textit{partial zero dynamics} surface, then these outputs, when properly controlled on the exoskeleton, yield stable periodic walking. The orbit is \textit{impact-invariant} if it returns to the partial zero dynamics surface $\mathcal{PZ}_{\alpha}$ after an impact event. To find the polynomials $\alpha$ that yield an impact-invariant periodic orbit on the reduced-order manifold, we formulate an optimization problem of the form: 
\begin{align}
    \alpha^{\ast} = \underset{\alpha}{\operatorname{argmin}}\qquad &\mathcal{J}(\alpha), \label{eqn:optimization} \\
    \textrm{s.t.} \qquad &\Delta(\mathcal{S}\cap \mathcal{PZ}_{\alpha}) \subset \mathcal{PZ_{\alpha}}, \label{eq:HZD}\\
    & \mathcal{W}_ix \leq b_i, \label{eq:constraints}
\end{align}
where $\mathcal{J}(\alpha)$ is a user-determined cost, \eqref{eq:HZD} is the impact invariance condition, \eqref{eq:constraints} are other physical constraints, $\mathcal{S}$ is the \textit{guard} defining the conditions under which impulsive behavior occurs, and $\Delta$ is the \textit{reset map} governing the system's dynamical response to hitting the guard. 

The optimization in \eqref{eqn:optimization}-\eqref{eq:constraints} produces a gait that can be altered by varying the cost function $\mathcal{J}(\alpha)$ and/or adding physical constraints. In bipedal walking, this cost is frequently the mechanical cost of transport (COT) defined by Eqs. (17)-(18) in \cite{reher2016algorithmic}. To create the desired motion, one must add physical constraints such as step length and foot height.

\newsec{Gait Generation Applied to Lower-Body Exoskeletons} To translate gait generation to lower-body exoskeletons, one must choose the optimization cost function and physical constraints to obtain user-preferred gaits. While it is possible to optimize generated gaits for mechanical properties such as COT, there is currently no well-understood relationship between the parameters of the optimization problem and user preferences. Additionally, due to the time-consuming nature of gait generation---both the time required to tune the optimization problem's constraints and the time required to run the program---the issue of generating human-preferred dynamically-stable walking gaits remains largely unexplored.

\newsec{Gait Library} It has become increasingly common to pre-compute a set of nominal walking gaits over a grid of various parameters \cite{da2017supervised}. These pre-computed gaits are combined to form a ``gait library,'' through which gaits can be selected and executed immediately. For the purpose of exoskeleton walking, a gait library allows the operator to select a gait that is comfortable for the patient; however, it is not yet clear how to select an appropriate walking gait to optimize user comfort and preference. Thus, we consider learning from the user's preferences, as discussed below.

%% file: Sections/Algorithm.tex
\section{Preference-based Learning Algorithm}
\label{sec:algorithm}
We leverage {\em preference-based learning} (e.g., does the user prefer gait A over gait B?) to determine the gait parameters most preferred by the user \cite{sui2018advancements,yue2012k,shivaswamy2015coactive,furnkranz2010preference,dorsa2017active,furnkranz2012preference}, since preference feedback has been shown to be much more reliable than absolute feedback when learning from subjective human responses \cite{sui2018advancements,joachims2005accurately}.  Thus, our goal to personalize the exoskeleton's gait can be framed as {\em dueling bandit}  \cite{sui2018advancements, yue2012k} and {\em coactive learning} \cite{shivaswamy2012online, shivaswamy2015coactive} problems.


Our work builds upon the Self-Sparring algorithm, a Bayesian dueling bandits approach that enjoys both competitive theoretical convergence guarantees and empirical performance \cite{sui2017multi}. 
Self-Sparring learns a Bayesian posterior over each action's utility to the user and draws multiple samples from the model's posterior to ``duel'' or ``spar'' via preference elicitation. The Self-Sparring algorithm iteratively: a) draws multiple samples from the posterior model of the actions' utilities; b) for each sampled model, executes the action with the highest sampled utility; c) queries for preference feedback between the executed actions; and d) updates the posterior according to the acquired preference data.

To collect more feedback beyond just one bit per preference, we also 
allow the user to suggest improvements during their trials. This approach resembles the \textit{coactive learning} framework \cite{shivaswamy2012online, shivaswamy2015coactive}, 
in which the user identifies an improved action as feedback to each presented action. Coactive learning has been applied to robot trajectory planning \cite{jain2015learning, somers2016human}, but has not, to our knowledge, yet been applied to robotic gait generation or in concert with preference learning. 

\newsec{The \algo~Algorithm} To optimize an exoskeleton's gait within the gait library (Section \ref{sec:stable_gaits}), we propose the \algo~algorithm, a \textit{mixed-initiative} learning approach \cite{wolfman2001mixed, lester1999lifelike} which extends the Self-Sparring algorithm to incorporate coactive feedback. Similarly to Self-Sparring, \algo~maintains a Bayesian \textit{preference relation function} over the possible actions, which is fitted to observed preference feedback. 
\algo~updates this model with user feedback and uses it to select actions for new trials and to elicit feedback. We first define the Bayesian preference model, and then detail the steps of Algorithm \ref{alg:coactive_self_sparring}.

\newsubsec{Modeling Utilities from Preference Data} 
We adopt the preference-based Gaussian process model of \cite{chu2005preference}.
Let $\mathcal{A} \subset \mathbb{R}^d$ be the finite set of available actions with cardinality $A = |\mathcal{A}|$. 
At any point in time, \algo~has collected a preference feedback dataset $D = \{\bm{x}_{k1} \succ \bm{x}_{k2} \, | \, k = 1, \ldots, N\}$ consisting of $N$ preferences, where $\bm{x}_{k1} \succ \bm{x}_{k2}$ indicates that the user prefers action $\bm{x}_{k1}\in\mathcal{A}$ to action $\bm{x}_{k2}\in\mathcal{A}$ in preference $k$. Furthermore, we assume each action $\bm{x}_i$ has a latent, underlying utility to the user, $f(\bm{x}_i)$. For finite action spaces, the utilities can be written in vector form: $\bm{f} := [f(\bm{x}_1), f(\bm{x}_2), \ldots, f(\bm{x}_A)]^T$. Given preference data $D$, we are interested in the posterior probability of $\bm{f}$:
\begin{equation}\label{eqn:posterior}
P(\bm{f} | D) \propto P(D | \bm{f})P(\bm{f}).
\end{equation}
We define a Gaussian prior over $\bm{f}$:

\begin{equation*}
P(\bm{f}) = \frac{1}{(2\pi)^{A/2} |\Sigma|^{1/2}} \text{exp}\left(-\frac{1}{2} \bm{f}^T \Sigma^{-1} \bm{f}\right),
\end{equation*}
where $\Sigma \in \mathbb{R}^{A \times A}$, $[\Sigma]_{ij} = \mathcal{K}(\bm{x}_i, \bm{x}_j)$, and $\mathcal{K}$ is a kernel, for instance the squared exponential kernel.

For computing the likelihood $P(D | \bm{f})$, we assume  feedback may be corrupted by i.i.d. Gaussian noise:  when presented with action $\bm{x}_i$, the user determines  her internal valuation $y(\bm{x}_i) = f(\bm{x}_i) + \varepsilon_i$, where $\varepsilon_i \sim \mathcal{N}(0, \sigma^2)$. Then,
\begin{flalign*}
P(\bm{x}_{k1} \succ \bm{x}_{k2} | \bm{f}) &= P(y(\bm{x}_{k1}) > y(\bm{x}_{k2}) | f(\bm{x}_{k1}), f(\bm{x}_{k2})) \\ &= \Phi\left[\frac{f(\bm{x}_{k1}) - f(\bm{x}_{k2})}{\sqrt{2}\sigma} \right],
\end{flalign*}
where $\Phi$ is the standard normal cumulative distribution function, and $y(\bm{x}_{kj}) = f(\bm{x}_{kj}) + \varepsilon_{kj}$, $j \in \{1, 2\}$. Thus, the full expression for the likelihood is:
\begin{equation}
    P(D | \bm{f}) = \prod_{k = 1}^N \Phi\left[\frac{f(\bm{x}_{k1}) - f(\bm{x}_{k2})}{\sqrt{2}\sigma} \right]. \label{eqn:likelihood}
\end{equation}

The posterior $P(\bm{f} | D)$ can be estimated via the Laplace approximation as a multivariate Gaussian distribution; see \cite{chu2005preference} for details. Finally, in formulating the posterior, preferences can be weighted relatively to one another if some are thought to be noisier than others. This is accomplished by changing $\sigma$ to $\sigma_k$ in \eqref{eqn:likelihood} to model differing values of the preference noise parameter among the data points, and is analogous to weighted Gaussian process regression \cite{hong2017weighted}.

\begin{algorithm}[t]
\caption{\algo}
\begin{small}
\begin{algorithmic}[1]
\Procedure{\algo}{$\mathcal{A}$ = action set, $n$ = number of actions to select at each iteration, $b$ = buffer size, $(\Sigma, \sigma)$ = utility prior parameters, $\beta$ = coactive feedback weight}
\State $D = \emptyset$ \Comment{Initialize preference dataset}
\State Obtain prior $(\bm{\mu}_0, \Sigma_0)$ over $\mathcal{A}$ from $(\Sigma, \sigma)$ \label{lin:learn_prior}
\ForAll{$t=1,2,\ldots $}
    \ForAll{$j = 1, \ldots, n$} \label{lin:sample_begin}
        \State Sample utility function $f_j$ from $(\bm{\mu}_{t - 1}, \Sigma_{t - 1})$
        \State Select action $a_j(t) = \text{argmax}_{x \in \mathcal{A}} f_j(x)$ \label{lin:select_action}
    \EndFor \label{lin:sample_end}
    \State Execute $n$ actions; observe pairwise feedback matrix $R = \{r_{jk} \in \{0, 1, \emptyset\}\}_{n \times (n + b)}$ \label{lin:execute}
    \ForAll{$j = 1, \ldots, n; k = 1, \ldots, n + b$}
      \If{$r_{jk}\neq\emptyset$}
        \State Append preference to dataset $D$
      \EndIf
    \EndFor
    \ForAll{$j = 1, \ldots, n$}
        \State Obtain coactive feedback $\tilde{a}_j(t) \in \mathcal{A} \cup \emptyset$ \label{lin:coactive_feedback}
      \If{$\tilde{a}_j(t) \neq\emptyset$}
        \State Add to $D$: $\tilde{a}_j(t)$ preferred to $a_j(t)$, weight $\beta$
      \EndIf        
    \EndFor 
    \State Update Bayesian posterior over $D$ to obtain $(\bm{\mu}_t, \Sigma_t)$ \label{lin:learn_post}
\EndFor
\EndProcedure
\end{algorithmic}
\end{small}
\label{alg:coactive_self_sparring}
\end{algorithm}

\newsubsec{The Learning Algorithm} Let $(\Sigma, \sigma)$ represent the prior parameters of the Bayesian preference model, as outlined above. From these parameters, one obtains the prior mean and covariance, $(\bm{\mu}_0, \Sigma_0)$ (Line \ref{lin:learn_prior} in Alg. \ref{alg:coactive_self_sparring}). In each iteration, \algo~updates the utility model (Line \ref{lin:learn_post}) via the Laplace approximation to the posterior in \eqref{eqn:posterior} to obtain $\mathcal{N}(\bm{\mu}_t, \Sigma_t)$.

\begin{figure*}[t!]
    \centering
    \includegraphics[width = \linewidth]{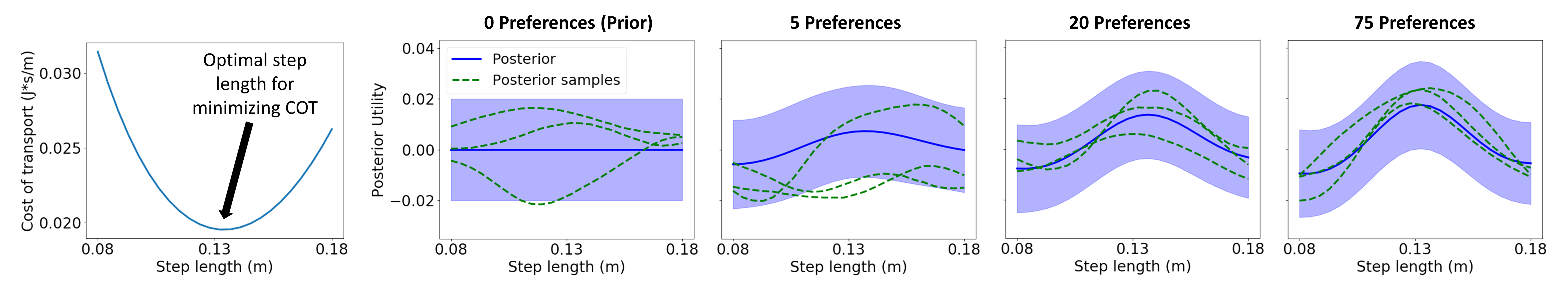}
    \vspace{-0.05in}
    \caption{Leftmost: COT for the CG biped at different step lengths and a fixed 0.2 m/s velocity. Remaining plots: posterior utility estimates of \algo~($n = 2$, $b = 0$; without coactive feedback) after varying iterations of learning (posterior mean +/- 2 standard deviations). The plots each show 3 posterior samples, which lie in the high-confidence region (mean +/- 2 stds) with high probability. The posterior utility estimate quickly converges to identifying the optimal action.}
    \label{fig:CompassBiped}
\end{figure*}

To select actions in the $t$\textsuperscript{th} iteration (Lines \ref{lin:sample_begin}-\ref{lin:sample_end}), the algorithm first draws $n$ samples from the posterior, $\mathcal{N}(\bm{\mu}_{t - 1}, \Sigma_{t - 1})$. Each of these is a utility function $f_j$, giving a utility value for each action in $\mathcal{A}$. The corresponding selected action is simply the one maximizing $f_j$ (Line \ref{lin:select_action}).

The $n$ actions are executed (Line \ref{lin:execute}), and the user provides pairwise preference feedback between pairs of actions (the user can always state ``no preference''). We extend Self-Sparring \cite{sui2017multi} to extract more preference comparisons from the available trials by assuming that the user can remember the $b$ actions \textit{preceding} the current $n$ actions. The user thus provides preferences between any combination of the current $n$ actions and the previous $b$ actions. 
For instance, for $n = 1$, $b > 0$, one can interpret $b$ as a buffer of previous trials that the user remembers. 
For $n = b = 1$, the user can report preferences between any pair of two consecutive trials, i.e., the user is asked, ``Did you like this trial more or less than the last trial?'' 
We expect that setting $n = 1$ while increasing $b$ to as many trials as the user can accurately remember would minimize the trials required to reach a preferred gait. In Line \ref{lin:execute}, the pairwise preferences from iteration $t$ form a matrix $R \in \mathbb{R}^{n \times (n + b)}$, where $r_{jk} \in \{0, 1, \emptyset\}$; the values $0$ and $1$ express preference information, while $\emptyset$ denotes the lack of a preference between the actions concerned. 

Finally, the user can suggest improvements in the form of coactive feedback (Line \ref{lin:coactive_feedback}). For example, the user could request a longer or shorter step length. In Line \ref{lin:coactive_feedback}, $\emptyset$ indicates that no coactive feedback was provided. Otherwise, the user's suggestion is appended to the data $D$ as preferred to the previously-tested action. In learning the model posterior, one can assign the coactive preferences a smaller weight relative to pairwise preferences via the input parameter $\beta > 0$.

\begin{figure}[t]
    \centering
    \subfloat[Objective function\label{fig:simTrue}]{ 
    \includegraphics[width = 0.5\linewidth]{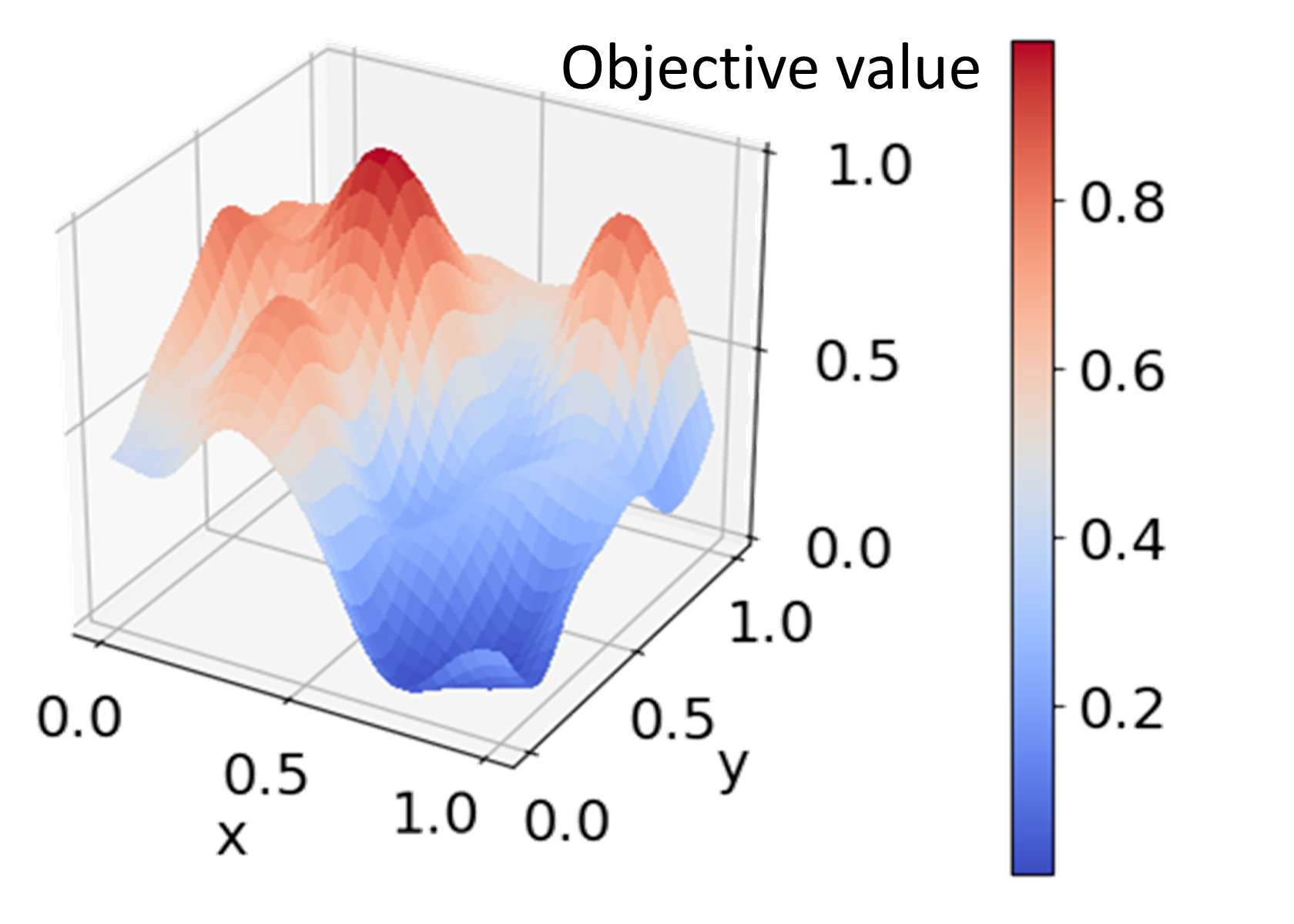}}
    \subfloat[Model posterior\label{fig:simPosterior}]{ 
    \includegraphics[width = 0.5\linewidth]{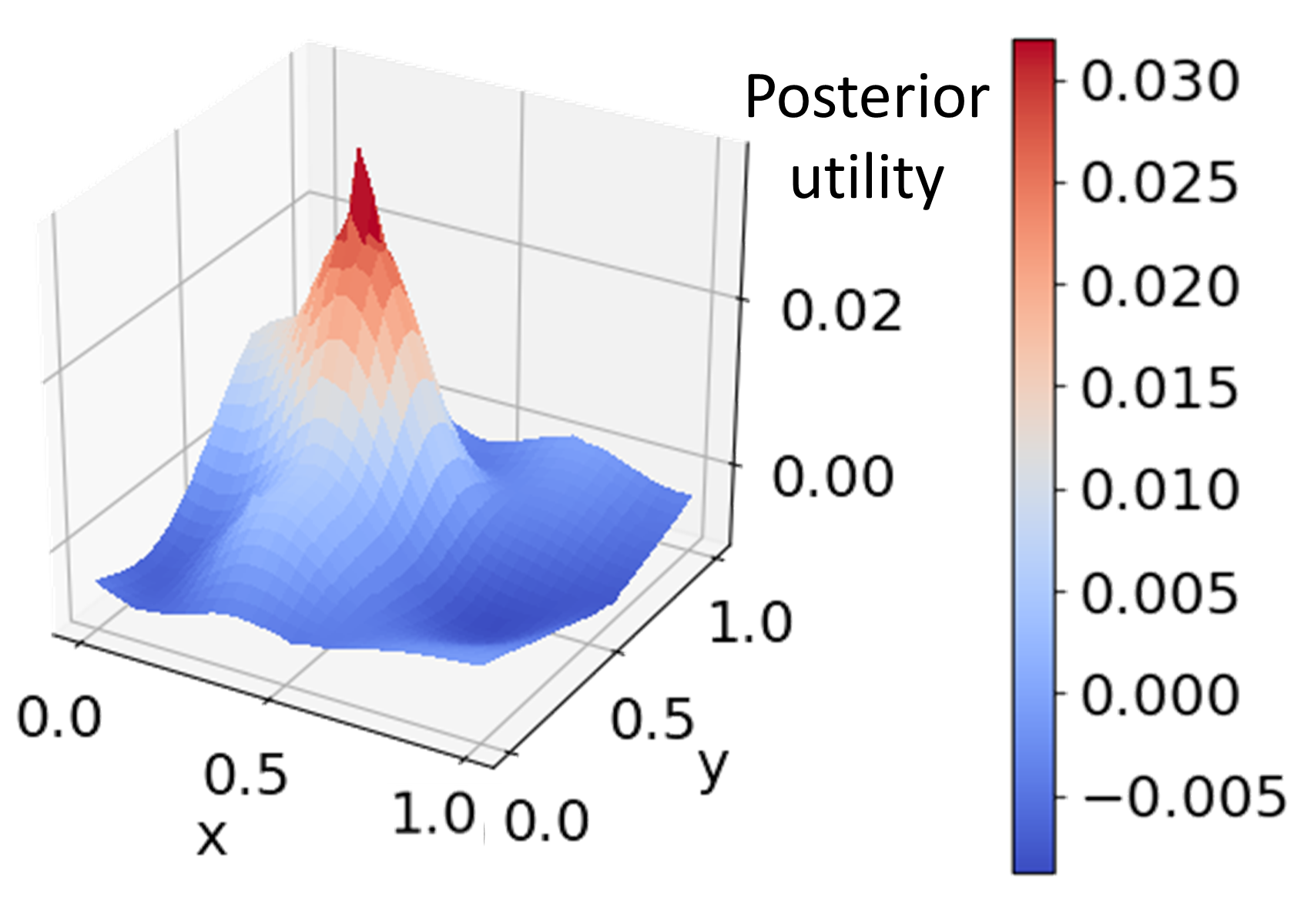}}
    \caption{a) Example synthetic 2D objective function. b) Utility model posterior learned after 150 iterations of \algo~ in simulation ($n = 1$; $b = 1$; coactive feedback). \algo~prioritizes identifying and exploring the optimal region, rather than learning a globally-accurate utility landscape.  }
\end{figure}


%% file: Sections/Simulation.tex
\section{Simulation Results}
The performance of \algo~is evaluated in two sets of simulations: (1) the compass-gait (CG) biped's COT,\footnote{Bayesian model's kernel: squared exponential with lengthscale = 0.025, signal variance = 0.0001, noise variance = 1e-8; preference noise ($\sigma$) = 0.01} and (2) a set of synthetic optimization objective functions.\footnote{Kernel: squared exponential with lengthscale = $[0.15, 0.15]$, signal variance = 0.0001, noise variance = 1e-5; preference noise ($\sigma$) = 0.01} In both cases, \algo~efficiently converges to the optimum.

\newsec{Optimizing the Compass-Gait Biped's Cost of Transport} We first evaluate our approach with a simulated CG biped, optimizing its COT over the step length via preference feedback (Fig. \ref{fig:CompassBiped}). Preferences are determined by comparing COT values, calculated by simulating gaits for multiple step lengths, each at a fixed forward hip velocity of 0.2 m/s. These simulated gaits were synthesized via a single-point shooting partial hybrid zero dynamics method \cite{westervelt2018feedback}.


We use \algo~to minimize the one-dimensional objective function in Fig. \ref{fig:CompassBiped}, using pairwise preferences obtained by comparing COT values for the different step lengths. Here, we use \algo~ with $n = 2$, $b = 0$, and without coactive feedback. Note that without a buffer or coactive feedback, \algo~reduces to  Self-Sparring \cite{sui2017multi}. At each iteration, two new samples are drawn from the Bayesian posterior, and the resultant two step lengths are compared to elicit a preference. Using the new preferences, \algo~updates its posterior over the utility of each step length.

Fig. \ref{fig:CompassBiped} depicts the evolution of the posterior preference model, where each iteration corresponds to a preference between two new trials. With more preference data, the posterior utility increasingly peaks at the point of lowest COT.  These results suggest that \algo~can efficiently identify high-utility actions from preference feedback alone.

\begin{figure}[t]
    \centering
    \includegraphics[width = \linewidth]{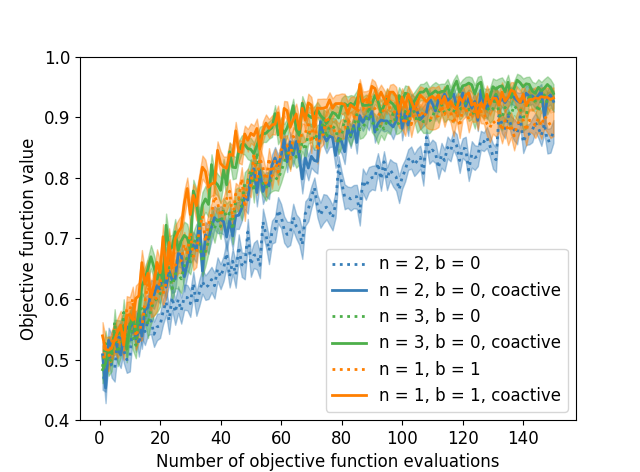}
    \caption{\algo~simulation results on 2D synthetic objective functions, comparing \algo~with and without coactive feedback for three parameter settings $n$ and $b$ (see Algorithm \ref{alg:coactive_self_sparring}). Mean +/- standard error of the objective values achieved over 100 repetitions. 
    The maximal and minimal objective function values are normalized to 0 and 1. We see that coactive feedback always helps, and that $n = 2$, $b = 0$---which receives the fewest preferences---performs worst.}
    \label{fig:sims_2D}
\end{figure}

\begin{figure*}[t]
    \centering
    \includegraphics[width = \linewidth]{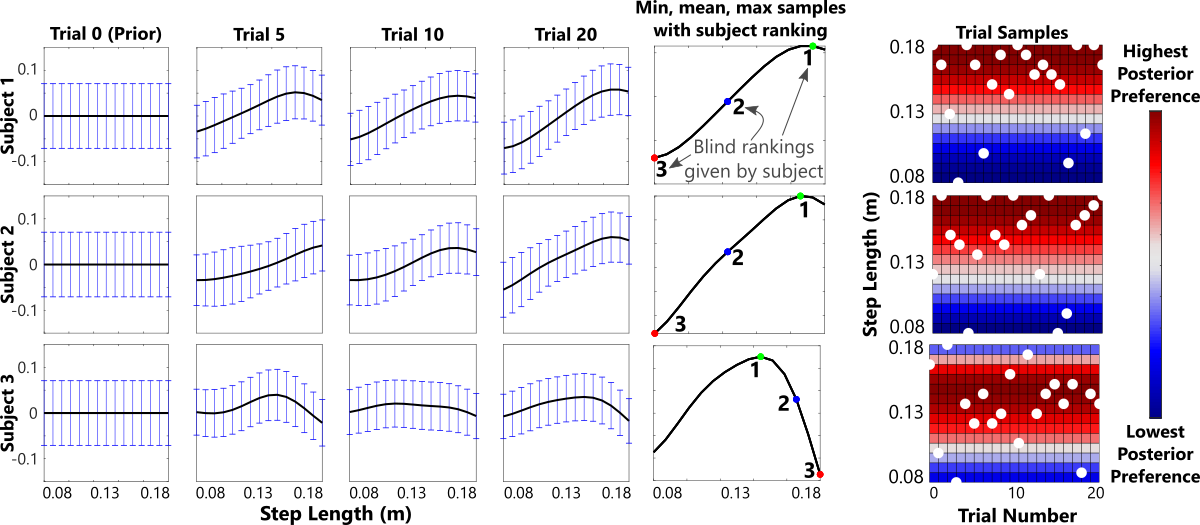}
    \caption{Experimental results for optimizing step length with three subjects (one row per subject). Columns 1-4 illustrate the evolution of the preference model posterior (mean +/- standard deviation), shown at various trials. \algo~converges to similar but distinct optimal gaits for different subjects. Column 5 depicts the subjects' blind ranking of the 3 gaits sampled after 20 trials. The rightmost column displays the experimental trials in chronological order, with the background depicting the posterior preference mean at each step length. \algo~draws more samples in the region of higher posterior preference.}
    \label{fig:OneFeaturePosterior}
\end{figure*}

\newsec{Optimizing Synthetic Two-Dimensional Functions} We next test \algo~ on synthetic 2D utility functions, such as the one shown in Fig. \ref{fig:simTrue}. Each utility function was generated from a Gaussian process prior on a 30-by-30 grid.  These experiments evaluate the potential to scale \algo~to higher dimensions and the advantages of coactive feedback.

We compare three settings for \algo's $(n,b)$ parameters: $(2,0)$, $(3,0)$, $(1,1)$ as explained in Sec. \ref{sec:algorithm}. For each setting---as well as with and without coactive feedback---we simulate \algo~on each of the 100 random objective functions. In each case, the number of objective function evaluations, or experimental trials, was held constant at 150.

Coactive feedback is simulated using a 2nd-order differencing approximation of the objective function's gradient. 
If \algo~selects a point at which both gradient components have magnitudes below their respective 50\textsuperscript{th} percentile thresholds, then no coactive feedback is given. Otherwise, we consider the higher-magnitude gradient component, and depending on the highest threshold that it exceeds (50\textsuperscript{th} or 75\textsuperscript{th}), simulate coactive feedback as either a $5\%$ or $10\%$ increase in the appropriate direction and dimension. 


Fig. \ref{fig:sims_2D} shows the simulation results. In each case, the mixed-initiative simulations involving coactive feedback improve upon those with only preferences. Learning is slowest for $n = 2, b = 0$ (Fig. \ref{fig:sims_2D}), since that case elicits the fewest preferences. Fig. \ref{fig:simPosterior} depicts the utility model's posterior mean for the objective function in Fig. \ref{fig:simTrue}, learned in the simulation with $n = 1$, $b = 1$, and mixed-initiative feedback. In comparing Fig. \ref{fig:simPosterior} to Fig. \ref{fig:simTrue}, we see that \algo~learns a sharp peak around the optimum, as it is designed to converge to sampling preferred regions, rather than giving the user undesirable options by exploring elsewhere.

\begin{figure*}[t]
    \centering
    \includegraphics[width = \linewidth]{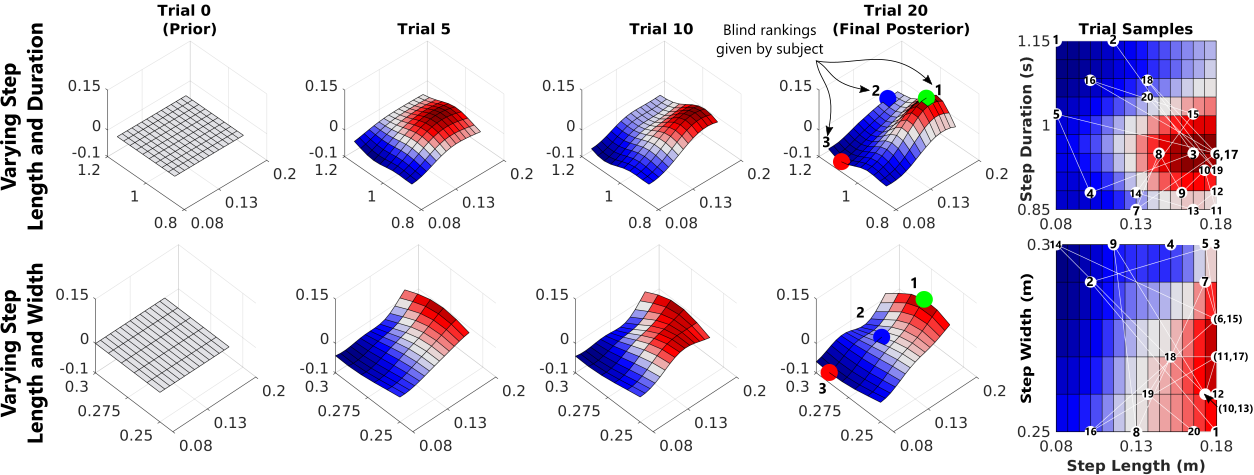}
    \caption{Experimental results from two-dimensional feature spaces (top row: step length and duration; bottom row: step length and width). Columns 1-4 illustrate the evolution of the preference model's posterior mean. Column 4 also shows the subject's blind ranking of the 3 gaits sampled after 20 trials. Column 5 depicts the experimental trials in chronological order, with the background as in Fig. \ref{fig:OneFeaturePosterior}. \algo~draws more samples in the region of higher posterior preference.} 
    \label{fig:TwoFeatures}
    \vspace{3pt}
\end{figure*}

\begin{figure*}[tb]
    \centering
    \includegraphics[width = \linewidth]{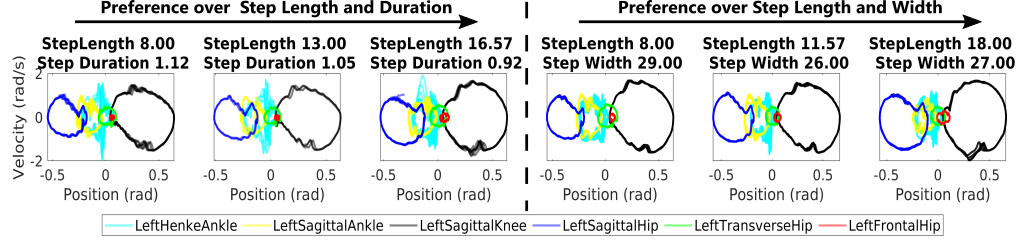}
    \caption{Experimental phase diagrams of the left leg joints over 10 seconds of walking. The gaits shown correspond to the maximum, mean, and minimum preference posterior values for both of subject 1's 2D experiments. For instance, subject 1 preferred gaits with longer step lengths, as shown by the larger range in sagittal hip angles in the phase diagram.}
    \label{fig:PhaseDiagrams}
\end{figure*}

%% file: Sections/Experiments.tex
\section{Human Subject Experiments}
After its validation in simulation, \algo~was deployed on a lower-body exoskeleton, Atalante, in two personalized gait optimization experiments with human subjects (video: \cite{video}). Both experiments aimed to determine gait parameter values that maximize user comfort, as captured by preference and coactive feedback. 
The first experiment,\footnote{Kernel: squared exponential with lengthscale = $0.03$, signal variance = 0.005, noise variance = 1e-7; preference noise ($\sigma$) = 0.02 \label{footnote:hyper_1D_exp}} repeated for three able-bodied subjects, used \algo~to determine the user's preferred step length, i.e., optimizing over a one-dimensional feature space. The second experiment\footnote{Same parameters as in \textsuperscript{\ref{footnote:hyper_1D_exp}} except for step duration lengthscale = 0.08 and step width lengthscale = 0.03} demonstrates \algo's effectiveness in two-dimensional feature spaces, and  optimizes simultaneously over two different gait feature pairs. Importantly, \algo~operates independently of the choice of gait features. The subjects' metabolic expenditure was also recorded via direct calorimetry as shown in Fig. \ref{fig:atalante}, but this data was uninformative of user preferences, as users are not required to expend effort toward walking.


\newsec{Learning Preferences between Step Lengths} In the first experiment, all three subjects walked inside the Atalante exoskeleton, with \algo~selecting the gaits. We considered 15 equally-spaced step lengths between 0.08 and 0.18 meters, each with a precomputed gait from the gait library. Feature discretization was based on users' ability to distinguish nearby values. The users decided when to end each trial, so as to be comfortable providing feedback. Since users have difficulty remembering more than two trials at once, we used \algo~with $n = 1$ and $b = 1$, which corresponds to asking the user to compare each current trial with the preceding one. Additionally, we query the user for coactive feedback: after each trial, the user can suggest a longer or shorter step length ($\pm$20\% of the range), a \textit{slightly} longer or shorter step length ($\pm$10\%), or no feedback. 

Each participant completed 20 gait trials, providing preference and coactive feedback after each trial. Fig. \ref{fig:OneFeaturePosterior} illustrates the posterior's evolution over the experiment. After only five exoskeleton trials, \algo~was already able to identify a relatively-compact preferred step length subregion. After the 20 trials, three points along the utility model's posterior mean were selected: the maximum, mean, and  minimum. The user walked in the exoskeleton with each of these step lengths in a randomized ordering, and gave a blind ranking of the three, as shown in Fig. \ref{fig:OneFeaturePosterior}. For each subject, the blind rankings match the preference posterior obtained by \algo, indicating effective learning of individual user preferences.

\newsec{Learning Preferences over Multiple Features} We further demonstrate \algo's practicality to personalize over multiple features, by optimizing over two different feature pairs: 1) step length and step duration and 2) step length and step width. The protocol of the 1D experiment was repeated for subject 1, with step lengths discretized as before, step duration discretized into 10 equally-spaced values between 0.85 and 1.15 seconds (with 10\% and 20\% modifications under coactive feedback), and step width into 6 values between 0.25 and 0.30 meters (20\% and 40\%). After each trial, the user was queried for both a pairwise preference and coactive feedback. Fig. \ref{fig:TwoFeatures} shows the results for both feature spaces. The estimated preference values were consistent with a 3-sample blind ranking evaluation, suggesting that \algo~successfully identified user-preferred parameters. Fig. \ref{fig:PhaseDiagrams} displays phase diagrams of the gaits with minimum, mean, and maximum posterior utility values to illustrate the difference between preferred and non-preferred gaits. 

%% file: Sections/Conclusion.tex
\section{CONCLUSIONS}
This work develops and demonstrates (video: \cite{video}) the \algo~interactive learning framework for optimizing gaits with respect to user comfort, using human preferences as feedback. We demonstrate the algorithm in simulation, showing that it efficiently learns to select optimal actions. We next apply \algo~in a user study with the Atalante lower-body exoskeleton, demonstrating the first application of preference-based learning for optimizing dynamic crutchless walking. \algo~successfully models the users' preferences, identifying compact subregions of preferred gaits.

In the future, we plan to apply \algo~toward optimizing over larger sets of gait parameters; this will likely require integrating the algorithm with techniques for learning over high-dimensional feature spaces \cite{kirschner2019adaptive}. The method could also be extended beyond working with precomputed gait libraries to generating entirely new gaits  or controller designs (e.g., via preference-based reinforcement learning \cite{furnkranz2012preference, novoseller2019dueling}).